\documentclass[review]{elsarticle}
\usepackage{geometry}
\usepackage{lineno,hyperref}
\usepackage{booktabs}
\usepackage{graphicx}
\usepackage{multicol}
\usepackage{adjustbox}
\usepackage{multirow}
\usepackage{caption}
\usepackage{amsmath}
\usepackage{subcaption}
\usepackage{color}
\usepackage{fancyhdr}
\usepackage{array}
\usepackage[flushleft]{threeparttable}
\usepackage{rotating}
\usepackage[withpage]{acronym}
\usepackage{float}
\usepackage{algorithm} 
\usepackage{lineno}
\usepackage{algpseudocode} 
\usepackage{verbatim} 
\usepackage{amsfonts}
\usepackage{ulem}
\usepackage{dirtytalk}
\modulolinenumbers[5]
\usepackage{natbib} 
\usepackage{multirow}
\usepackage{hyperref}
\journal{Journal}
\date{}

\begin{document}
\title{Automated Road Crack Localization to Guide Highway Maintenance}
\begin{frontmatter}

\author[1,2,3]{Steffen Knoblauch \corref{cor1}}
\author[2]{Ram Kumar Muthusamy}
\author[4,5]{Pedram Ghamisi}
\author[1,2,3]{Alexander Zipf}
\cortext[cor1]{Corresponding author: steffen.knoblauch@uni-heidelberg.de}
\address[1]{HeiGIT at Heidelberg University, Heidelberg, Germany}
\address[2]{GIScience Research Group, Heidelberg University, Heidelberg, Germany}
\address[3]{Interdisciplinary Centre of Scientific Computing (IWR), Heidelberg University, Heidelberg, Germany}
\address[4]{Lancaster University, Bailrigg, LA1 4YR Lancaster, United Kingdom}
\address[5]{Helmholtz-Zentrum Dresden-Rossendorf (HZDR), Helmholtz Institute Freiberg for Resource Technology, Freiberg, Germany}

\begin{abstract}
Highway networks are crucial for economic prosperity. Climate change-induced temperature fluctuations are exacerbating stress on road pavements, resulting in elevated maintenance costs. This underscores the need for targeted and efficient maintenance strategies. This study investigates the potential of open-source data to guide highway infrastructure maintenance. The proposed framework integrates airborne imagery and OpenStreetMap (OSM) to fine-tune YOLOv11 for highway crack localization. To demonstrate the framework's real-world applicability, a Swiss Relative Highway Crack Density (RHCD) index was calculated to inform nationwide highway maintenance. The crack classification model achieved an F1-score of 0.84 for the positive class (crack) and 0.97 for the negative class (no crack). The Swiss RHCD index exhibited weak correlations with Long-term Land Surface Temperature Amplitudes (LT-LST-A) (Pearson’s \(r = -0.05\)) and Traffic Volume (TV) (Pearson’s \(r\) = 0.17), underlining the added value of this novel index for guiding maintenance over other data. Significantly high RHCD values were observed near urban centers and intersections, providing contextual validation for the predictions. These findings highlight the value of open-source data sharing to drive innovation, ultimately enabling more efficient solutions in the public sector.
\end{abstract}


\end{frontmatter}


\newpage
\section{Introduction\label{chap:introduction}}
Critical infrastructure forms the backbone of modern society, encompassing essential systems and assets such as transportation networks (e.g., roads, bridges, and railways), energy grids (e.g., electricity and gas supply), water distribution systems, and communication networks. These infrastructures are fundamental to economic stability, public safety, and national security, as their disruption can lead to cascading failures that affect multiple sectors \cite{di_pietro_critical_2021}. Among these, road infrastructure is particularly crucial, providing the foundation for mobility, facilitating the transportation of goods and people, and supporting economic activities  \cite{Amador-Jimenez01062012, verburg_global_2011}. Highways, as integral components of the road network, enable high-capacity, long-distance travel and serve as key corridors connecting major urban centers, industrial zones, and logistic hubs \cite{percoco_highways_2016}. Their efficiency and reliability are essential for national and regional connectivity, making them a critical focus for infrastructure planning and resilience strategies \cite{vijayakumar_social_2024}. To ensure that highways serve their purpose effectively, they must be maintained regularly, as external factors such as high TV, temperature fluctuations, and heavy axle loads impose significant stress on the pavement structure \cite{fifer_bizjak_impact_2014, sen_effect_2022}. Without proper maintenance, these factors can lead to deterioration, reducing the safety and efficiency of highway networks, and requiring costly repairs \cite{rioja2013infrastructure}.

Effective highway maintenance is essential for preserving infrastructure quality and ensuring road safety. Traditionally, maintenance decisions have been guided by reactive approaches such as damage reports, traffic volume analysis, and accident records \cite{borghetti_road_2024}. While these methods provide valuable insights, they are inherently limited by their dependence on manual inspections and retrospective data, which may result in delayed interventions and increased maintenance costs. As a result, research has increasingly focused on leveraging advanced computational techniques to improve the accuracy and efficiency of road condition assessment \cite{kumar_pavement_2024, saeed_review_2020}. Recent advancements in remote sensing and computer vision have introduced automated methods for monitoring pavement conditions. Several benchmark datasets, such as DeepCrack \cite{liu_deepcrack_2019} and RDD \cite{arya_rdd2022_2022, arya_global_2020}, have accelerated the development and evaluation of deep-learning models for crack detection and classification.


Numerous studies have explored diverse model architectures and distress types. For instance, \citet{mandal_deep_2020} systematically compared backbone networks such as CSPDarknet53, Hourglass-104, and EfficientNet, contributing to a better understanding of trade-offs in model performance. Broader distress categorization—covering longitudinal, transverse, and alligator cracks, potholes, delamination, and repairs—has been addressed in multi-label detection frameworks \cite{zhang_road_2022, zhong_multi-scale_2022, samadzadegan_automatic_2024}. \citet{zhong_multi-scale_2022} introduced PDDNet, a multitask fusion model that integrates region-level and pixel-level detection, while \citet{kang_hybrid_2020} extended detection to include physical crack measurements using a modified tubularity flow field (TuFF) algorithm and distance transform method (DTM). Several studies aim to overcome data scarcity or improve robustness. \citet{zhong_deeper_2023} applied GAN-based augmentation to better train models on grooved concrete pavement imagery. \citet{haciefendioglu_concrete_2022} evaluated model stability under varying environmental conditions, including lighting and weather, which is critical for real-world deployment. Lightweight YOLO-based models - such as DenseSPH-YOLOv5 \cite{roy_densesph-yolov5_2023}, YOLOv5-CBoT \cite{yu_improved_2023}, YOLO-LRDD \cite{wan_yolo-lrdd_2022}, and YOLOv8-PD \cite{zeng_yolov8-pd_2024} - have been proposed for real-time applications, particularly those suited for mobile or edge devices. Despite these advancements, large-scale operational deployment remains a key bottleneck. Most existing models are tested on small-scale settings or lab-controlled benchmark datasets. \citet{zhao_highway_2023} is among the few to attempt a scaled application, using UAV imagery and a CrackNet-based approach over 108 km of highways. \citet{bashar_exploring_2022} explored satellite imagery for crack detection, offering a broader spatial perspective but with lower resolution trade-offs. These efforts highlight the potential for scalable monitoring, yet the large-scale deployment of such methods in operational highway maintenance remains limited, with public sector practices still largely reliant on manual reporting and visual inspections.

This study makes a substantive contribution toward closing the gap between academic research and practical implementation in highway maintenance. Leveraging openly available high-resolution airborne imagery and pretrained convolutional neural networks - specifically YOLOv11 - it presents an automated, scalable framework for precise crack localization on highways, capable of being routinely re-executed as new imagery becomes available. A key innovation is the Road Highway Crack Detection (RHCD) index, a novel metric designed to quantify pavement distress using remote sensing data. The framework is evaluated through a case study in Switzerland, addressing two central research questions: (RQ1) To what extent can pavement cracks on highways be detected at scale using airborne imagery? and (RQ2) To what extent do long-term land surface temperature anomalies (LT-LST-A) and traffic volume (TV) correlate with the RHCD index? The proposed approach is both flexible and transferable, offering the potential for broader application across diverse regional contexts. These contributions support the development of data-driven maintenance strategies in the public sector, advancing the transition toward a more proactive and resilient model of infrastructure management.

\section{Materials and Methods}\label{materials_and_methods}
\begin{figure}[htbp]
  \centering
  \includegraphics[width=0.9\linewidth]{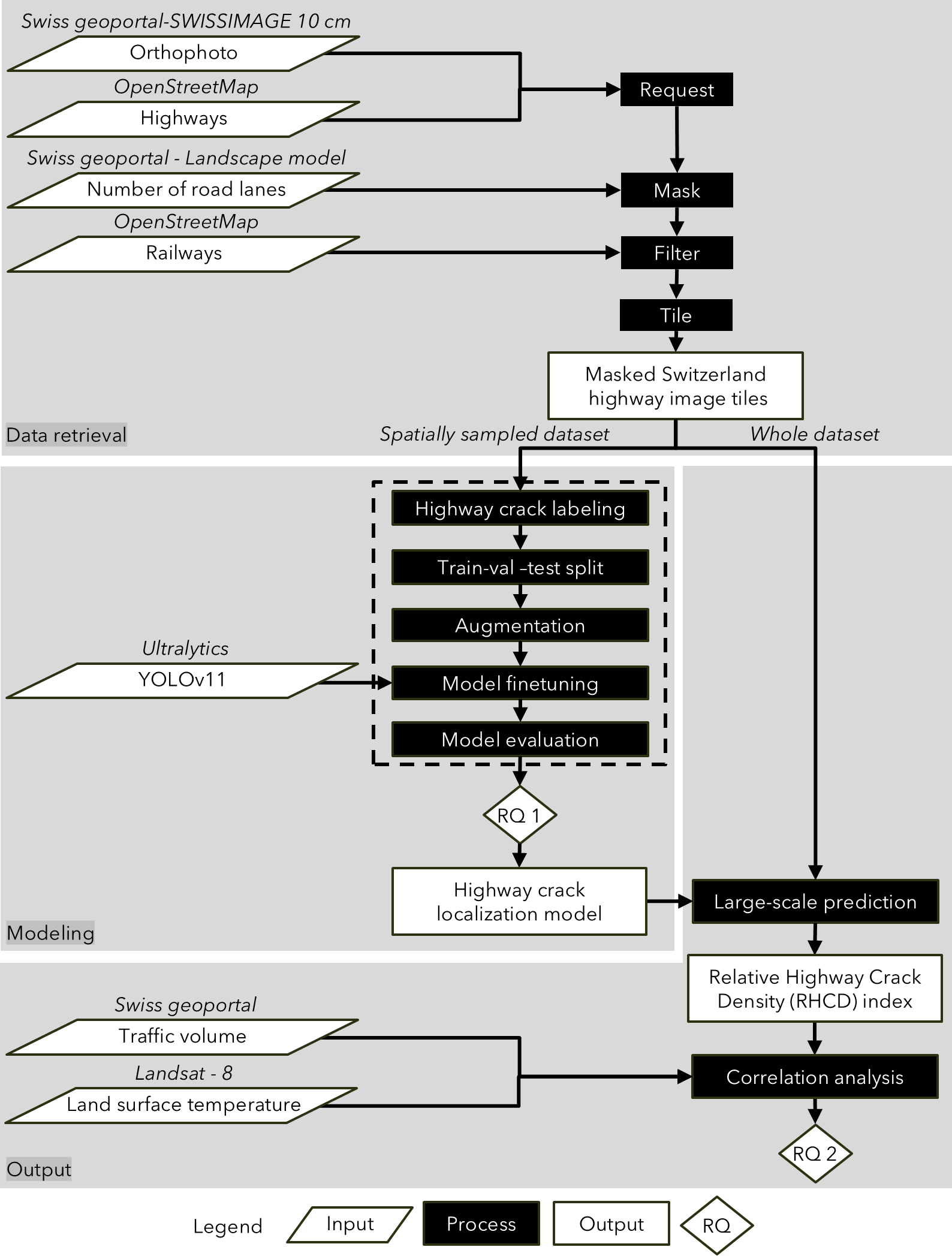}
  \vspace{0.5cm}
  \caption{Schematic representation of the proposed framework for scalable highway crack detection, comprising i) the retrieval of road imagery, ii) the training of a crack localization model, and iii) the generation of a RHCD index.} 
  \label{fig:workflow_chart} 
\end{figure}

This study introduces a scalable and data-driven framework for highway crack localization, aimed at supporting road infrastructure maintenance strategies at a national level. As illustrated in Figure \ref{fig:workflow_chart}, the proposed approach comprises three main components: (1) acquisition and preprocessing of geospatial imagery - here, from Switzerland - where orthophotos are (i) retrieved via an API using OpenStreetMap highway vectors excluding tunnels, (ii) masked to exclude non-road surfaces, and (iii) filtered to remove segments obscured by overpasses such as railway bridges; (2) development of a crack localization model trained on a spatially stratified sample of image tiles, involving annotation, data augmentation, transfer learning, and performance evaluation, followed by large-scale inference across the entire highway network; and (3) aggregation of model predictions into a Relative Highway Crack Density (RHCD) index, which is subsequently analyzed in relation to TV and LT-LST-A.

\subsection{Data}
This study draws on four openly accessible datasets (see Table~\ref{tab:datasets}). Highway imagery and transportation network data serve as the foundation for constructing the RHCD index. To contextualize the index and illustrate its added value, additional datasets - specifically TV and LT-LST-A data - are integrated into the analysis. A detailed map of the study area is provided in \ref{AppendixA}.

\begin{table}[H]
\centering
\scriptsize
\caption{Overview of datasets used in this study}
\label{tab:datasets}
\begin{tabular}{|p{3.2cm}|p{3.2cm}|p{2.2cm}|p{2.2cm}|p{2.5cm}|}
\hline
\textbf{Data} & \textbf{Source} & \textbf{Coverage \newline Period} & \textbf{Geotype} & \textbf{Resolution} \\ \hline

Highway Imagery & Swiss Federal Office of Topography (swisstopo) \newline  via Swiss Geoportal & 2021–2023 & Raster & 10 cm (lowlands), \newline  25 cm (Alps) \\ \hline

Transportation \newline  Infrastructure Network & OpenStreetMap via Geofabrik + swissTLMRegio & September 2024 & Vector (polyline) & -- \\ \hline

Traffic Volume (TV) & Federal Office for Spatial Development (ARE) \newline  via Swiss Geoportal & 2017 & Vector (polyline) & -- \\ \hline

Land Surface \newline  Temperature (LST) & USGS Landsat 8 \newline  via Google Earth Engine & 2021–2023 & Raster & 30 m \\ \hline

\end{tabular}
\end{table}

\subsubsection{Highway Imagery}
As openly-accessible highway imagery we picked the SWISSIMAGE 10 cm orthophoto mosaic covering the entirety of Switzerland. The dataset offers a ground resolution of 10 cm in the lowlands and main Alpine valleys, with a 25 cm resolution in the Alps, and is provided by the Swiss Federal Office of Topography (swisstopo) \cite{swissttopo}. Each image tile is provided in Cloud-Optimized GeoTIFF (COG) format, featuring RGB color bands, and is updated on a 3-year cycle, making it highly suitable for any detailed spatial analysis. The dataset comprises approximately 42,700 imagery tiles across Switzerland, each representing a 1 km × 1 km area, and is accessible through the Swiss Geoportal API \cite{swiss_stac_api}.

\subsubsection{Transportation Infrastructure Network}
As of 2024, the Swiss road network spans 85,009 km, including 2,259 km of national roads, of which 1,549 km are highways \cite{bfs2024infrastructure}. In this study transportation infrastructure network data - containing road and railway networks - were extracted from OpenStreetMap (OSM) in September 2024 via Geofabrik \cite{Geofabrik_2024}, a provider of free, preprocessed OSM extracts. OSM is a collaborative, open-source geographic database where features like roads and rails are represented as \texttt{ways} (polylines) with key-value tags. The completeness of the Swiss highway network was assessed and found to be high in September 2024 using the Ohsome Quality Analyst (OQT) \cite{noauthor_ohsome_nodate}. The data contains a total of 1,680,721 road segments, of which 8,081 correspond to highways, and 52,874 to railway tracks. Additional road lane information was derived from the swissTLMRegio Landscape Model (2024) \cite{swissttopo, swiss_stac_api} and matched with the OSM data.

\subsubsection{Traffic Volume}
Although highways constitute less than 3\% of the total Swiss road network, they accommodate approximately 45\% of vehicle-kilometers, emphasizing their critical role in sustaining high-capacity transport both within Switzerland and across Central Europe \cite{swissinfo2024traffic}. The Swiss road network facilitates substantial transport volumes, with 93.06 billion passenger-kilometers recorded in 2022 and an estimated 17.4 billion tonne-kilometers of road freight transported in 2021 \cite{oecd2022trends, statista2021freight}. In 2024, highway congestion increased by 22.4\% compared to the previous year, reflecting escalating demand that also accelerates pavement degradation \cite{swissinfo2024traffic}. The increase was particularly notable on the A1 and A2 highways, which are critical corridors for European supply chains, serving as vital links between major economic hubs within the European 'Blue Banana' economic zone. The TV data utilized in this study was obtained from the Federal Office for Spatial Development (ARE), Switzerland \cite{are_federal_office}. This dataset provides average daily traffic for the year 2017, covering a total of 260,825 road segments across the country. The reported TV ranges from 0 vehicles per day on low-traffic roads to a peak of 86,935 vehicles per day on the A1L highway in Zurich. The dataset was obtained through the Swiss Geoportal \cite{swiss_stac_api}.

\subsubsection{Land Surface Temperature}
Switzerland experiences high climatic variability, characterized by extreme seasonal temperature fluctuations, including summer heatwaves and winter cold spells \cite{fischer_climate_2022, meteoswiss2025slippery}. These variations impose considerable stress on critical infrastructure, particularly the national highway network, by accelerating material fatigue and pavement deterioration. The LST dataset used in this study is derived from the Landsat 8 Collection 2 Level-2 Science Product, provided by the U.S. Geological Survey \cite{usgs_landsat8_lst}. This product was accessed via Google Earth Engine (GEE) \cite{gorelick2017google}. We derived monthly bi-temporal LST images spanning from 2021 to 2023, ensuring temporal consistency with the captured highway imagery period. LST represents the radiative skin temperature of the land surface, influenced by factors such as solar radiation, surface materials, and atmospheric conditions \cite{weng_estimation_2004}.

\subsection{Methodology}
\subsubsection{Data retrieval}
High-resolution SWISSIMAGE orthophoto tiles (10 cm spatial resolution) from 2021 to 2023 were accessed via the Swiss Spatial Temporal Asset Catalog (STAC) API, hosted at \textit{data.geo.admin.ch}, focusing on the most recent imagery intersecting with OSM highway segments. These segments were filtered using the \texttt{highway=motorway} key, and to ensure surface-level highways were analyzed, segments classified as tunnels (\texttt{tunnel=yes}) were excluded. Following data acquisition, buffer zones were generated around OSM road centerlines using lane count information from the swissTLMRegio Landscape Model \cite{swissttopo} and an assumed average lane width of 3.5 meters \cite{wolhuter_geometric_2015, casali_topological_2019}. These buffers delineated the extent of the road surface and were used to mask out non-road pixels from the orthophotos, isolating relevant areas for subsequent analysis. Subsequently highway segments occluded by railway bridges, were excluded to ensure that only unobstructed, visually accessible road surfaces were retained for subsequent analysis. The cropped images were subdivided into 50 × 50 pixel tiles, corresponding to a ground area of 5 m × 5 m based on the 10 cm per pixel resolution, resulting in masked tiles representing highway segments in Switzerland.

\begin{figure}[H]
  \centering
  \includegraphics[width=\linewidth]{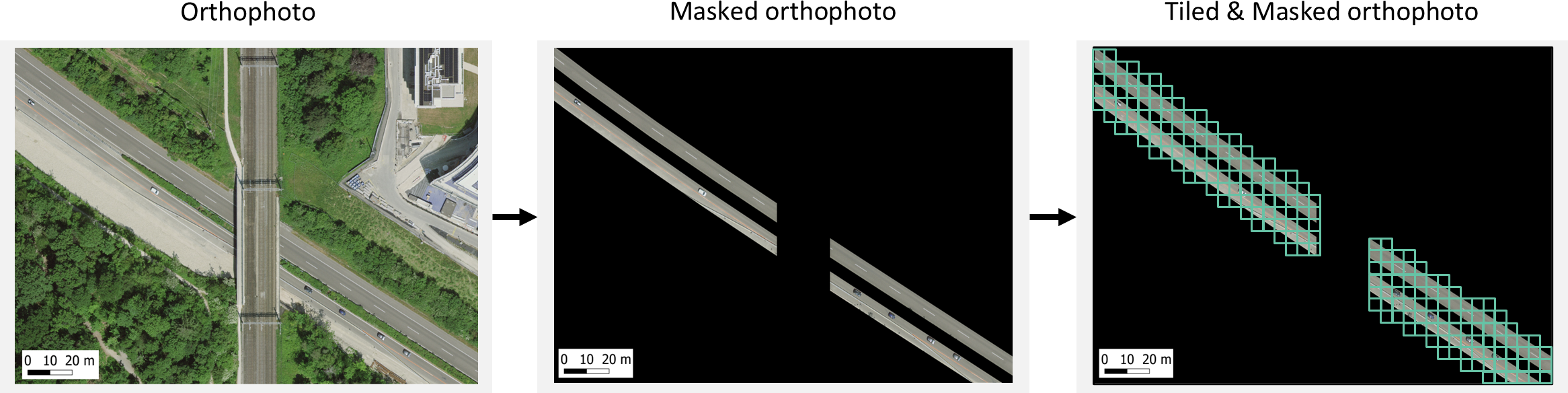}
  \caption{Overview of the data retrieval process: downloading, masking, and tiling of highway airborne imagery.}
  \label{fig:model}
\end{figure}

\subsubsection{Modeling}
The modeling pipeline commenced with the creation and manual annotation of a dataset comprising 31,556 image patches, including 4170 (13\%) samples exhibiting visible highway cracks and 27,386 (87\%) without detectable surface distress. The dataset was partitioned into training (80\%), validation (10\%), and test (10\%) subsets. To address the inherent class imbalance and enhance the model’s capacity to generalize under diverse environmental and structural conditions, extensive data augmentation was applied on the test set. This included geometric transformations (rotations of 90° and 270°, horizontal and vertical flipping) as well as photometric adjustments involving brightness modulation (±40 units). These operations were designed to simulate a range of illumination scenarios typically encountered in airborne imagery, such as shadows cast by adjacent vegetation, overpasses, or vehicles, and to capture variability in crack orientation. Following augmentation, the crack-positive test set was expanded sixfold to 23,345 samples, resulting in a final training dataset of  45,253 labeled patches.

\begin{table}[h]
\centering
\caption{Distribution of crack and no-crack annotations across dataset splits.}
\begin{tabular}{lccc}
\toprule
\textbf{Set} & \textbf{All Samples} & \textbf{Crack} & \textbf{No Crack} \\
\midrule
Train   & 45,253   & 23,345   & 21,908   \\
Validation  & 3,158   & 426   & 2,732    \\
Test & 3,156   & 417   & 2,739           \\
\bottomrule
\end{tabular}
\label{tab:crack_distribution}
\end{table}

Subsequently, a YOLOv11 deep learning architecture was fine-tuned using a focal loss function to mitigate the effects of class imbalance between crack-positive and crack-negative samples. This approach emphasizes harder-to-classify (i.e., crack-containing) image patches during training, thereby enhancing the model's ability to detect low-contrast or narrow cracking patterns that are often underrepresented in the training data. Model performance was evaluated through both quantitative and qualitative means. The F1 score was used to assess predictive accuracy on a held-out validation set, while visual inspections of activation patterns from Layer 8, a C3k2 module, were conducted using Guided Gradient-weighted Class Activation Mapping (Grad-CAM). These heatmaps facilitated interpretability by confirming that the model’s predictions were spatially aligned with features indicative of pavement cracking.

\begin{figure}[H]
  \centering
  \includegraphics[width=\linewidth]{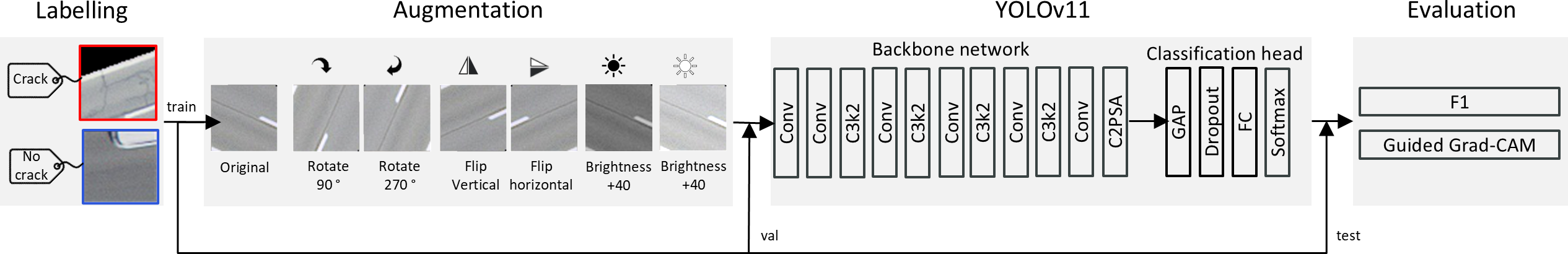}
  \caption{Overview of the modeling pipeline, including data labeling, augmentation of the training set, YOLOv11 model architecture design, and computation of evaluation metrics.}
  \label{fig:model}
\end{figure}

\subsubsection{Output}  
To translate highway crack localizations into actionable insights for infrastructure maintenance, we introduce a spatial index map representing the \textit{Road Highway Crack Density} (RHCD) across Switzerland. This aggregation reflects the practical reality that highway maintenance is typically conducted at the scale of extended road segments rather than at the fine-grained resolution (5 m × 5 m) used for initial crack localization (cf. \ref{AppendixB}). The RHCD for a given spatial unit \( i \in I \), where \( I \) denotes the set of all predefined spatial units within the study area, is defined as:

\begin{equation}
\text{RHCD}_i (\%) = \left( \frac{\text{Number of airborne imagery tiles containing cracks in } i}{\text{Total number of airborne imagery tiles in } i} \right) \times 100.
\label{eq:rhcd}
\end{equation}

This metric captures the proportion of image tiles within each spatial unit that contain detected pavement cracks, thereby providing a spatially explicit quantification of crack density. In this study, OSM road segments were selected as the spatial unit of analysis, owing to their accessibility, granularity, and alignment with real-world road infrastructure, including delineation of both driving directions. Subsequently, we assess the relationship between the RHCD index and two auxiliary variables: TV and LT-LST-A. These comparisons are visualized via scatterplots and quantified using Pearson's correlation coefficients. The LT-LST-A is computed at a 30\,m spatial resolution, using land surface temperature (LST) data aggregated over the period 2021–2023, and is defined as:

\begin{equation}
\text{LT-LST-A} = \max(\text{LST}) - \min(\text{LST}),
\label{eq:lt-lst-a}
\end{equation}
where \(\max(\text{LST})\) and \(\min(\text{LST})\) refer to the maximum and minimum raster values observed across the specified period. Traffic volume (TV) is evaluated over buffered road segments consistent with the spatial units used for RHCD calculation.

\section{Results and Discussion}\label{results_and_discussion}
Table \ref{tab:model_performance} summarizes the results of the highway crack localization model. For the positive class, the model achieved a precision of 0.81, indicating that 81\% of the predicted highway cracks were correct. The recall of 0.86 demonstrates that 86\% of the actual highway cracks were correctly identified. This resulted in an F1-score of 0.84, reflecting a well-balanced performance between precision and recall. Similarly, the model performed well for the negative class, with a precision of 0.98 and a recall of 0.97, yielding an F1-score of 0.97. These results suggest that the model effectively discriminates between positive and negative instances, with particularly high accuracy in identifying regions with no highway cracks while maintaining strong recall for positive instances.

\begin{table}[ht]
    \centering
    \begin{tabular}{|c|c|c|c|}
        \hline
        \textbf{Class} & \textbf{Precision} & \textbf{Recall} & \textbf{F1-score} \\
        \hline
        Highway Crack          & 0.81              & 0.86           & 0.84            \\
        No Highway Crack       & 0.98              & 0.97           & 0.97            \\
        \hline
    \end{tabular}
    \caption{Performance metrics of the fine-tuned model for highway crack localization.}
    \label{tab:model_performance}
\end{table}

Notably, the model demonstrated robustness across diverse lighting conditions and pavement textures due to both the augmentation strategy and the use of high-resolution input tiles (50×50 pixels). Ablation studies were performed to evaluate the contribution of key components such as data augmentation, weighted sampling, and model architecture selection. Removing the augmentation pipeline resulted in a 10.25\% drop in F1-score, highlighting the importance of synthetic variability in training. Replacing YOLOv11x-cls with standard YOLOv8 led to slower inference and a 5.15\% drop in accuracy, confirming the benefit of using the more advanced architecture. Furthermore, a test-time ensembling strategy combining predictions across augmented views improved classification consistency on borderline cases. In summary, the modeling pipeline effectively integrates modern deep learning practices with domain-specific enhancements, yielding a deployable crack localization system capable of scalable, automated highway condition assessment.

\begin{figure}[ht]
    \centering
    \includegraphics[width=\linewidth]{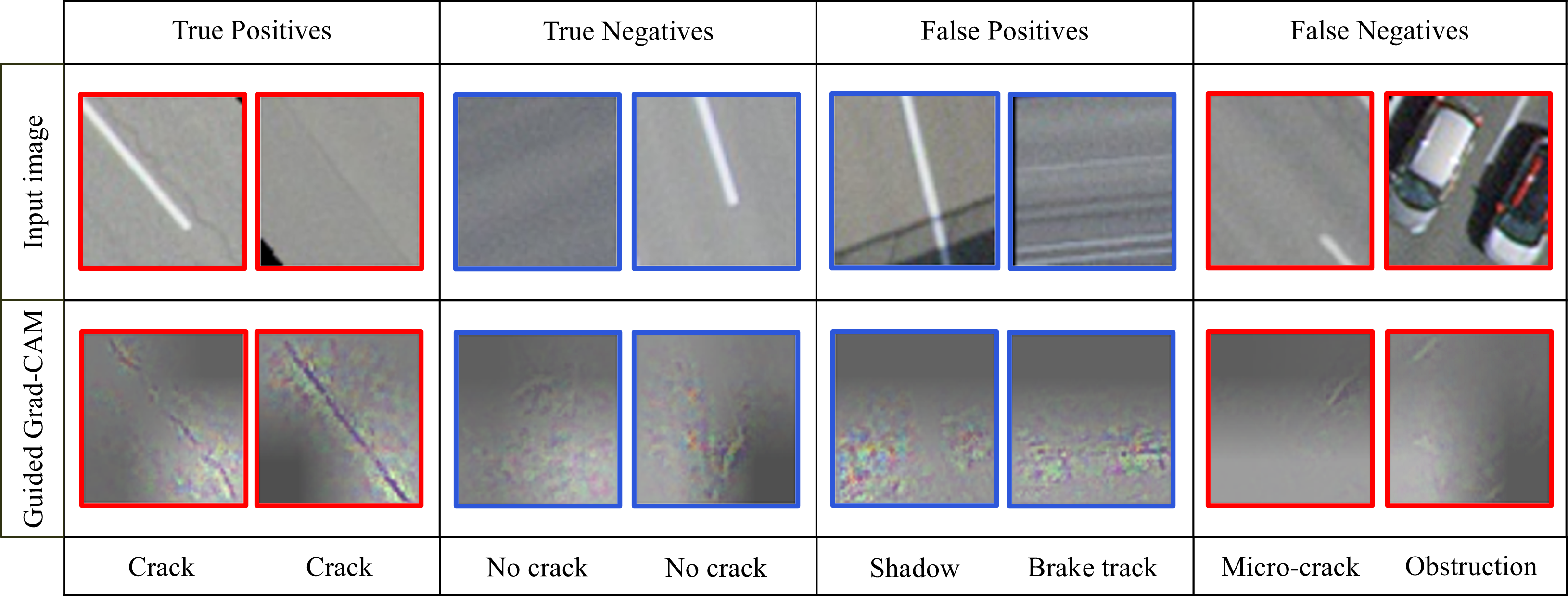}
    \caption{Visual inspection of highway crack localization results (red = positive detection; blue = negative detection). Two examples of each True Positives (TP), False Positives (FP), True Negatives (TN), and False Negatives (FN), with Guided Grad-CAM visualizations indicating key regions affecting predictions.}
    \label{fig:Visual_inspection}
\end{figure}

Figure \ref{fig:Visual_inspection} illustrates the classification performance with examples of True Positives (TP), False Positives (FP), True Negatives (TN), and False Negatives (FN), accompanied by Guided Grad-CAM visualizations, a technique that integrates Gradient-weighted Class Activation Mapping with guided backpropagation to emphasize critical image regions for mode prediction based on gradient-derived feature importance.

True Positive (TP) instances underscore the model’s capability to accurately detect highway cracks, even when lane markings are present. Similarly, True Negative (TN) instances demonstrate the model's effectiveness in correctly identifying regions devoid of highway cracks, thereby minimizing the occurrence of False Positives (FP). In the case of False Positives (FP), the model occasionally misclassifies non-crack features—such as shadows cast by bridges or light poles, and brake tracks—as highway cracks. This limitation reflects the model's difficulty in distinguishing visually similar patterns, particularly in aerial imagery with limited contextual information. False Negative (FN) instances, where highway cracks are not detected, typically arise from subtle micro-cracks or cracks obscured by vehicles during image acquisition. These challenges can impact the precision of the generated RHCD index.

To mitigate the number of False Positives, one potential approach is to incorporate these visually similar features as additional training classes. This would enable the model to more effectively differentiate between distinct yet visually similar structures, ultimately improving detection accuracy. Additionally, to address False Negatives, further model improvements could include advanced data augmentation techniques, such as adding noise, introducing lighting variations, or simulating occlusions. Exploring two-stage object detection models, such as Faster R-CNN, may also enhance detection accuracy by first generating region proposals and refining them before classification, which could provide more robust crack localization in challenging scenarios

\begin{figure}[ht]
    \centering
    \includegraphics[width=\linewidth]{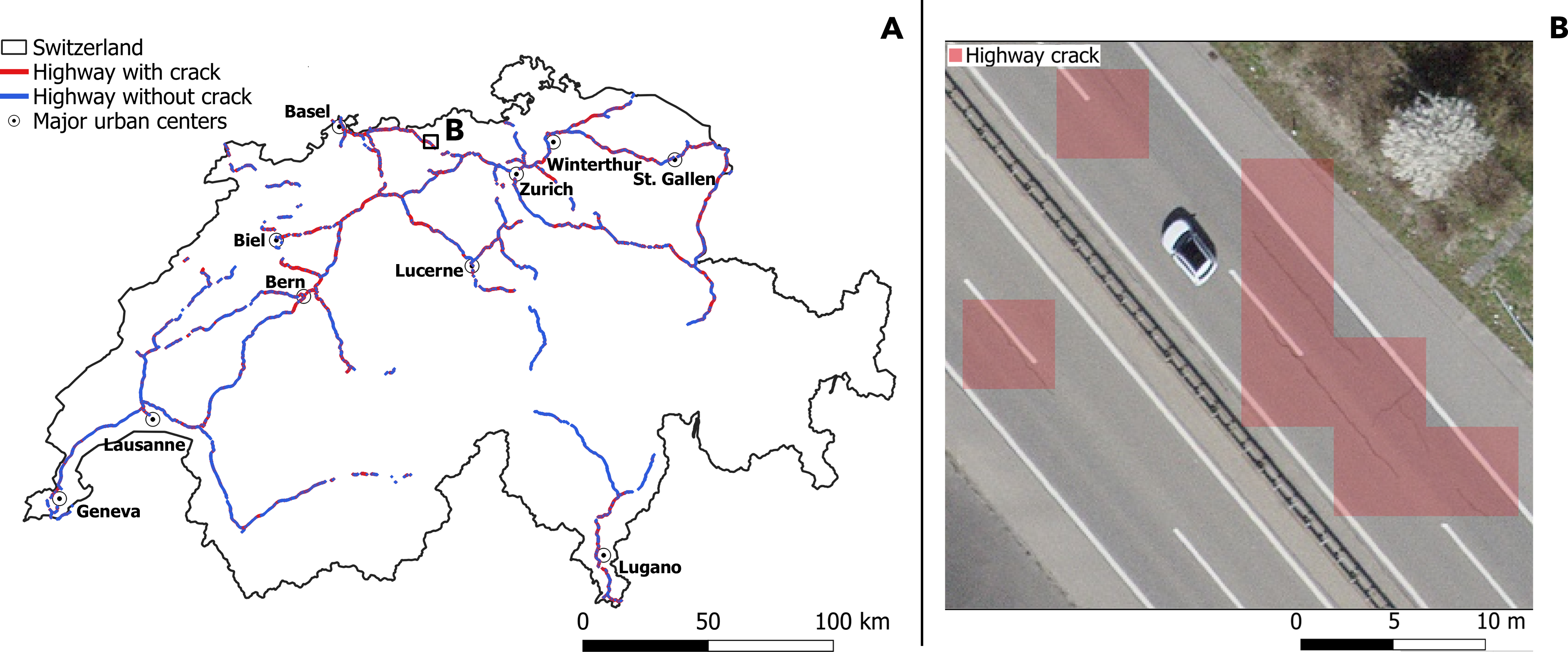}
    \caption{Panel A shows the Swiss highway crack localization map along with major urban centers in Switzerland and a bounding box indicating the area detailed in Panel B. Panel B presents a zoomed-in view highlighting the classification results on 5 m × 5 m airborne imagery tiles used for calculating the RHCD index.}
    \label{fig:crack_pred}
\end{figure}

Figure \ref{fig:crack_pred} illustrates the results of the large-scale crack localization across the Swiss highway network. Notably, elevated pavement stress was observed in proximity to urban centers and highway intersections, contrasting with segments in more remote Alpine regions. This spatial variation provides further contextual validation for the highway crack predictions.

\subsection{Large-scale Crack Localization}
\begin{figure}[H]
    \centering
    \includegraphics[width=\linewidth]{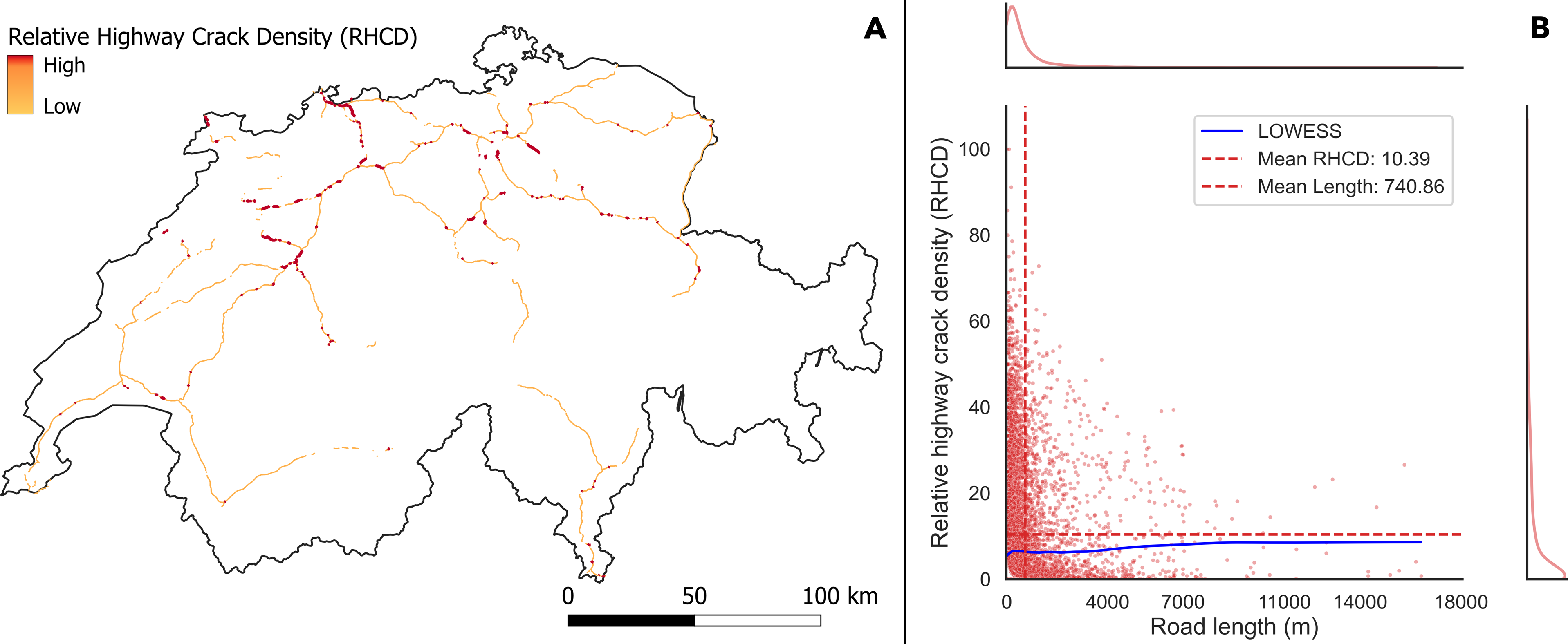}
    \caption{Panel A illustrates the spatial distribution of the RHCD index, calculated at the OSM highway segment level across Switzerland, with OSM road segments in the top 5\% of the RHCD index highlighted by thick red lines. Panel B illustrates the distribution of OSM road segment lengths in Switzerland relative to the estimated RHCD, emphasizing the influence of road segment length on index calculation despite the relative nature of the measure.}
    \label{fig:rhcd_map}
\end{figure}

Swiss highway maintenance is generally carried out at the road segment level, rather than at the 5 m x 5 m patch resolution used for crack localization. To address this, a RHCD index was developed at the OSM road segment level, as shown in Panel A of Figure \ref{fig:rhcd_map}. Panel B of Figure \ref{fig:rhcd_map} illustrates the influence of road segment length on the RHCD index calculation. Notably, there is a tendency for longer highway segments to exhibit lower index values, despite the relative nature of the measurement. 

\begin{figure}[H]
    \centering
    \includegraphics[width=\linewidth]{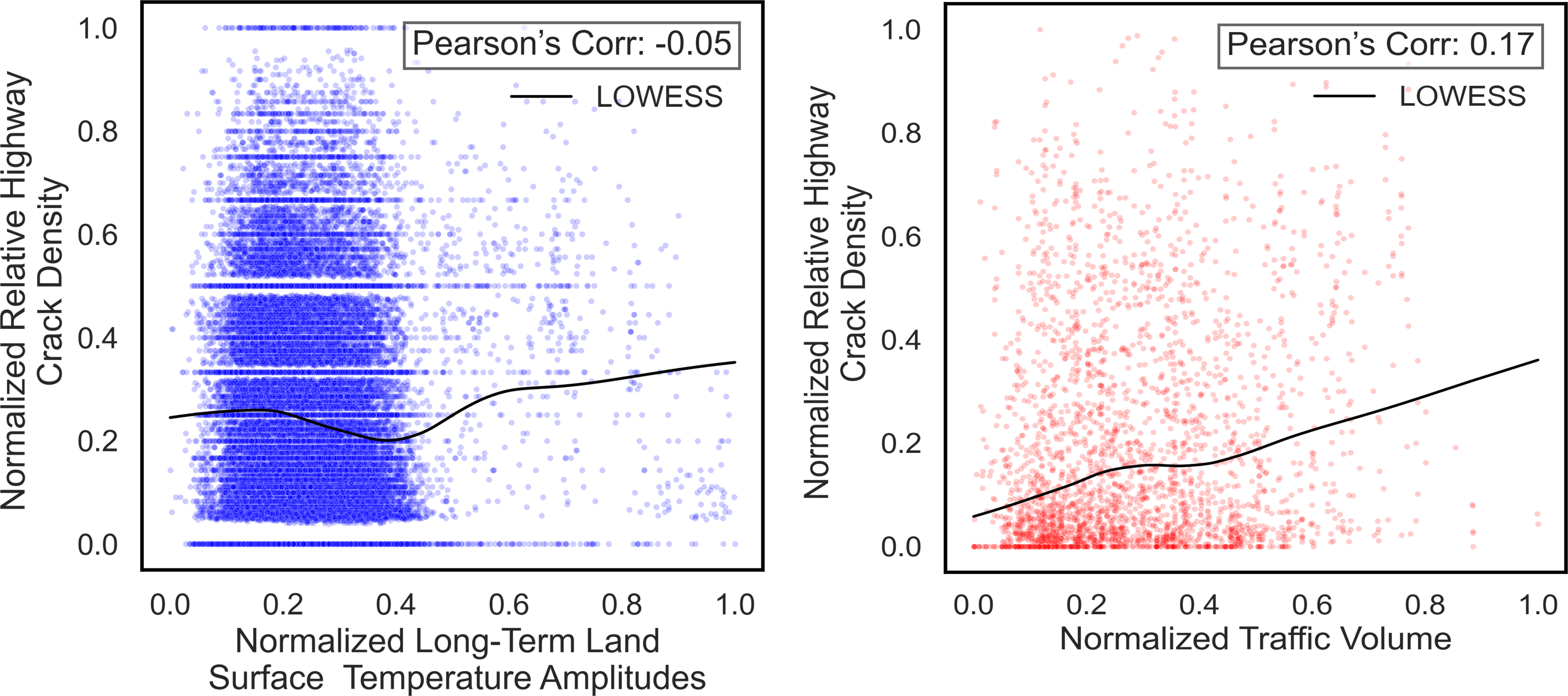}
    \caption{Comparison of normalized RHCD, detected from airborne imagery, with normalized LT-LST-A (left) and normalized TV (right), as alternative indicators for highway maintenance.}
    \label{fig:lst_traffic}
\end{figure}

Overlaying the RHCD index with additional data, such as the economic importance of roads, could further inform maintenance prioritization. The segment size used in the RHCD calculation can be adjusted within the workflow to any spatial unit, enabling customization to align with local maintenance strategies. Similarly, the RHCD index was compared with LT-LST-A measurements at a 30m grid resolution. Panel A in Figure \ref{fig:lst_traffic} shows a scatterplot of the RHCD and LT-LST-A comparison, indicating a very weak negative correlation (Pearson’s \(r\) = -0.05), suggesting little to no linear relationship between the two variables. The association between RHCD and TV was slightly stronger (Pearson’s \(r\) = 0.17), although still very weak. These low correlation values highlight the complexity of highway crack formation, suggesting that neither LT-LST-A nor TV alone serves as a strong predictor of RHCD. Crack formation is likely influenced by multiple factors simultaneously, and low correlation values do not necessarily imply the absence of a causal relationship, as such relationships may be non-linear, indirect, or obscured by unmeasured confounding variables. Confounding factors, such as pavement material and age, may contribute to the observed results. This complexity further emphasizes the value of imagery-based crack localization as a tool to guide highway maintenance decisions. Furthermore, the temporal mismatch of data - where LT-LST-A spans 2021-2023 and TV is from 2017 - could influence the findings, as the formation of highway cracks may have occurred over a longer time period.

\section{Conclusion}\label{conclusion}
In conclusion, this study highlights the potential of leveraging openly available high-resolution imagery and pretrained convolutional neural networks, such as YOLOv11, to achieve precise and scalable highway crack localization for informed infrastructure maintenance. The Swiss case study demonstrates the added value of the novel RHCD index when compared with other publicly accessible indicators, such as LT-LST-A and TV, highlighting its ability to provide more precise and actionable insights for highway maintenance. The proposed framework offers a flexible approach that can be applied across different regions with varying maintenance workflows. Future research could explore multivariate analyses to better understand the factors influencing highway crack occurrence, providing deeper insights and supporting the development of predictive maintenance solutions.

\section{Data availability}\label{Data availability}
The code for this analysis is publicly available at https://doi.org/10.5281/zenodo.15574877.

\newpage


\section*{Supplementary materials}\label{supplements}
\appendix
\renewcommand\thefigure{\thesection.\arabic{figure}}    
\setcounter{figure}{0}

\section{Study area}\label{AppendixA}
\begin{figure}[H]
  \centering
  \includegraphics[width=\linewidth]{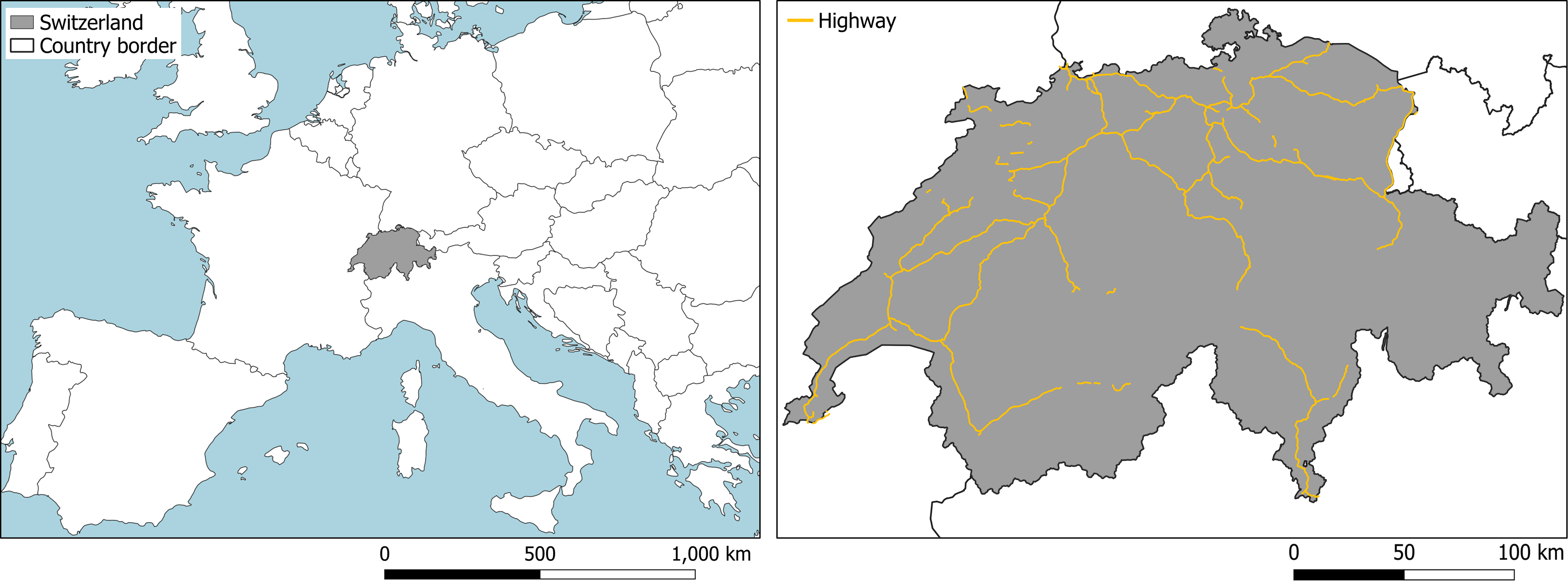}
  \caption{Map of Europe indicating the central geographic location of Switzerland (left) and map of highway segments without tunnel sections in Switzerland (right).}
  \label{fig:switzerland_highways}
\end{figure}

\section{Background on Swiss Highway Maintenance Guidance}\label{AppendixB}
\begin{figure}[H]
    \centering
    \includegraphics[width=0.7\linewidth]{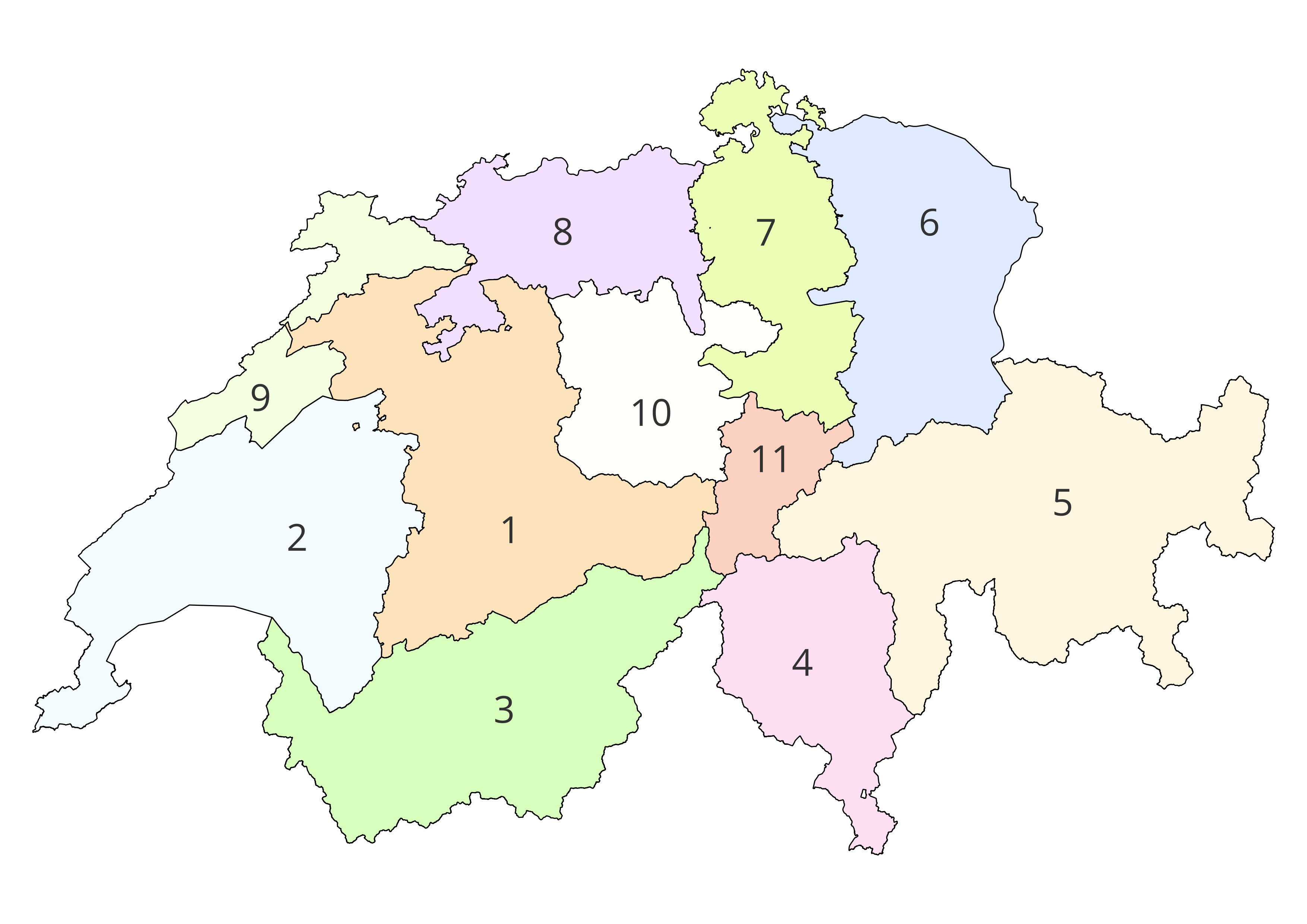}
    \caption{Map of territorial units responsible for managing the maintenance of Switzerland's national highway infrastructure. The units are organized into 11 regions under the oversight of the Federal Roads Office (FEDRO).}
    \label{fig:territory}
\end{figure}

The maintenance of Switzerland’s national highway network is a complex and resource-intensive process, coordinated through a structured system of funding and governance. In 2024, expenditures from the National Highways and Urban Transport Fund totaled CHF 3.399 billion, of which approximately CHF 1.737 billion was dedicated to the expansion and upkeep of national highways \cite{fedro_facts_nodate}. Operational responsibilities are shared between the Federal Roads Office (FEDRO) and the 26 cantons, which are organized into 11 territorial units (cf. Figure \ref{fig:territory}) tasked with managing the maintenance of national road infrastructure \cite{astra_gebietseinheiten_nodate}. These territorial units are embedded within cantonal construction departments and operate under service-level agreements with FEDRO. While FEDRO retains authority over traffic management and project oversight, implementation is delegated to these regional units \cite{astra_gebietseinheiten_nodate, efk_efficiency_nodate}.

Maintenance planning in Switzerland follows standardized federal guidelines designed to optimize resource allocation, ensure long-term infrastructure performance, and minimize disruption. Structural maintenance interventions are typically organized into 5-kilometer segments, spaced approximately 30 kilometers apart, with each segment expected to remain free of traffic-disrupting maintenance for at least 15 years after completion~\citep{astra_unterhaltsplanung_nodate}.

Traditionally, maintenance guidance is based on annual visual inspections conducted by territorial units~\cite{ruttimann_podcast_2024}. While effective, these inspections are labor-intensive and costly. To enhance operational efficiency, subcontractors such as STRABAG and Trimble have introduced mobile mapping technologies integrated into inspection vehicles, enabling faster and more systematic assessments~\cite{trimble_strabag_2024}. For civil engineering structures such as bridges and tunnels, additional sensor systems have been installed to monitor structural health. These sensors provide real-time data on critical parameters such as load-bearing capacity, vibration response, and material integrity, supporting early detection of degradation~\cite{kubadb}. A freely accessible, open-source platform for automated, large-scale condition monitoring of the national road network is still lacking. This study seeks to address this gap by illustrating the potential of open-source technologies and open data sharing to catalyze innovation in digital infrastructure maintenance.
\bibliography{crack_literature}

\end{document}